\documentclass[runningheads]{llncs}
\usepackage[T1]{fontenc}
\usepackage{xcolor}
\usepackage{enumitem}
\usepackage{fontawesome}
\usepackage{booktabs}
\usepackage{graphicx}
\usepackage{tabularx}
\usepackage{array, booktabs}
\usepackage{multirow}
\usepackage{makecell}
\usepackage{xcolor,colortbl}
\usepackage[most]{tcolorbox}
\usepackage[noend]{algpseudocode}
\usepackage{algorithmicx}
\usepackage{algorithm}
\usepackage{amsmath}
\usepackage[utf8]{inputenc} 
\usepackage[T1]{fontenc}      
\usepackage[pagebackref,breaklinks,colorlinks]{hyperref}
\usepackage{url}            
\usepackage{booktabs}       
\usepackage{amsfonts}       
\usepackage{nicefrac}       
\usepackage{microtype}      
\usepackage{float}
\usepackage{eufrak}
\usepackage{amsfonts}
\usepackage{yfonts}
\usepackage{amsmath}
\newcommand{\Xmat}[0]{{{\bf X}}}
\newcommand{\Hmat}[0]{{{\bf H}}}
\newcommand{\shortname}{WoLF}
\newcommand{\anatomy}{Anatomy-Specific Knowledge decoupling}
\newcommand{\shortanatomy}{ASK}
\definecolor{LightCyan}{rgb}{0.7,0.8,0.9}
\definecolor{Gray}{gray}{0.90}
\newcolumntype{g}{>{\columncolor{Gray}}c}

\begin{document}
\title{%
  \includegraphics[height=1em]{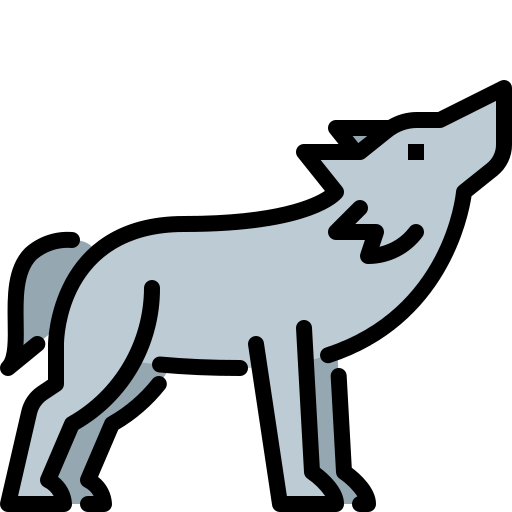} WoLF: Wide-scope Large Language Model Framework for CXR Understanding%
}
\titlerunning{WoLF: Large Language Model Framework for CXR Understanding}
\author{\small Seil Kang$^{1}$, Donghyun Kim$^{1}$, Junhyeok Kim$^{1}$,
Hyo Kyung Lee$^{2}$, Seong Jae Hwang$^{1}$}
\authorrunning{Kang et al.}
\institute{$^{1}$Yonsei University, $^{2}$Korea University \\
    \email{\{seil, danny0103, timespt, seongjae\}@yonsei.ac.kr,\\
    hyokyunglee@korea.ac.kr}
    }
\maketitle
\vspace{-0.5cm}
\begin{abstract}
Significant methodological strides have been made toward Chest X-ray (CXR) understanding via modern vision-language models (VLMs), demonstrating impressive Visual Question Answering (VQA) and CXR report generation abilities. However, existing CXR understanding frameworks still possess several procedural caveats. \underline{$\mathbf{\mathfrak{(1)}}$} Previous methods solely use CXR reports, which are insufficient for comprehensive Visual Question Answering (VQA), especially when additional health-related data like medication history and prior diagnoses are needed. \underline{$\mathbf{\mathfrak{(2)}}$} Previous methods use raw CXR reports, which are often arbitrarily structured. While modern language models can understand various text formats, restructuring reports for clearer, organized anatomy-based information could enhance their usefulness. \underline{$\mathbf{\mathfrak{(3)}}$} Current evaluation methods for CXR-VQA primarily emphasize linguistic correctness, lacking the capability to offer nuanced assessments of the generated answers. In this work, to address the aforementioned caveats, we introduce \textbf{WoLF}, a \textbf{W}ide-sc\textbf{o}pe \textbf{L}arge Language Model \textbf{F}ramework for CXR understanding. To resolve \underline{$\mathbf{\mathfrak{(1)}}$}, we capture multi-faceted records of patients, which are utilized for accurate diagnoses in real-world clinical scenarios. Specifically, we adopt the Electronic Health Records (EHR) to generate instruction-following data suited for CXR understanding. Regarding \underline{$\mathbf{\mathfrak{(2)}}$}, we enhance report generation performance by decoupling knowledge in CXR reports based on anatomical structure even within the attention step via masked attention. To address \underline{$\mathbf{\mathfrak{(3)}}$}, we introduce an AI-evaluation protocol optimized for assessing the capabilities of LLM. Through extensive experimental validations, WoLF demonstrates superior performance over other models on MIMIC-CXR in the AI-evaluation arena about VQA (up to  $+9.47\%$p mean score) and by metrics about report generation ($+7.3\%$p BLEU-1).
\end{abstract}
\section{Introduction}
\begin{figure}[!t]
    \includegraphics[width=\textwidth]{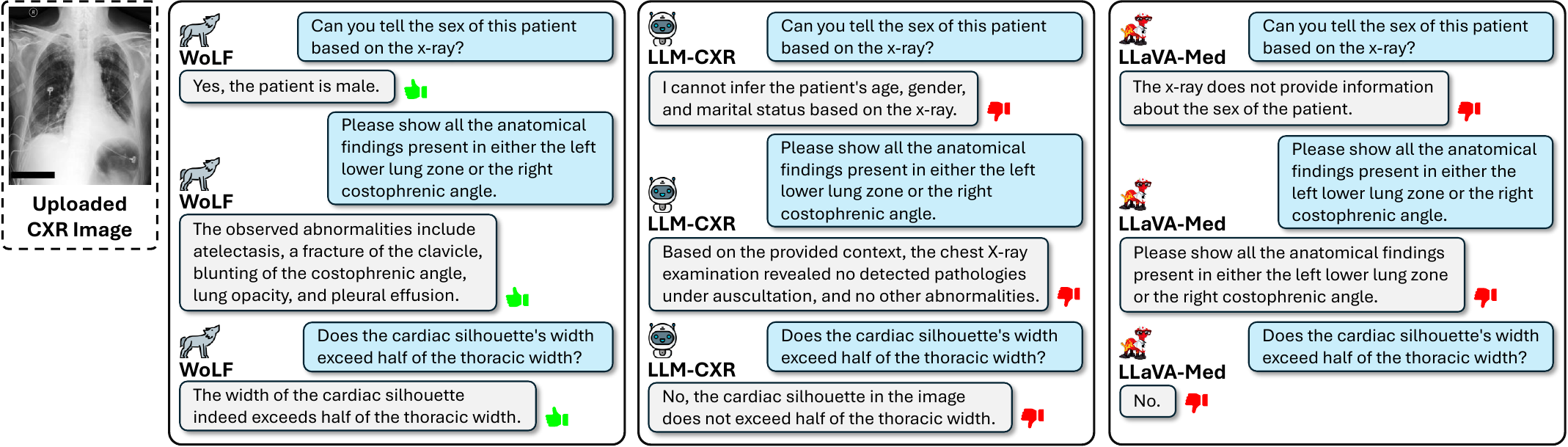}
\caption{Comparisons with other models for VQA scenario given a CXR image. Green thumbs indicate the quality of the response is good (accurate, helpful), while red thumbs indicate bad (inaccurate, evasive), with respect to target answers.} \label{fig1}
\end{figure}
\begin{figure}[t!]
    \includegraphics[width=\textwidth]{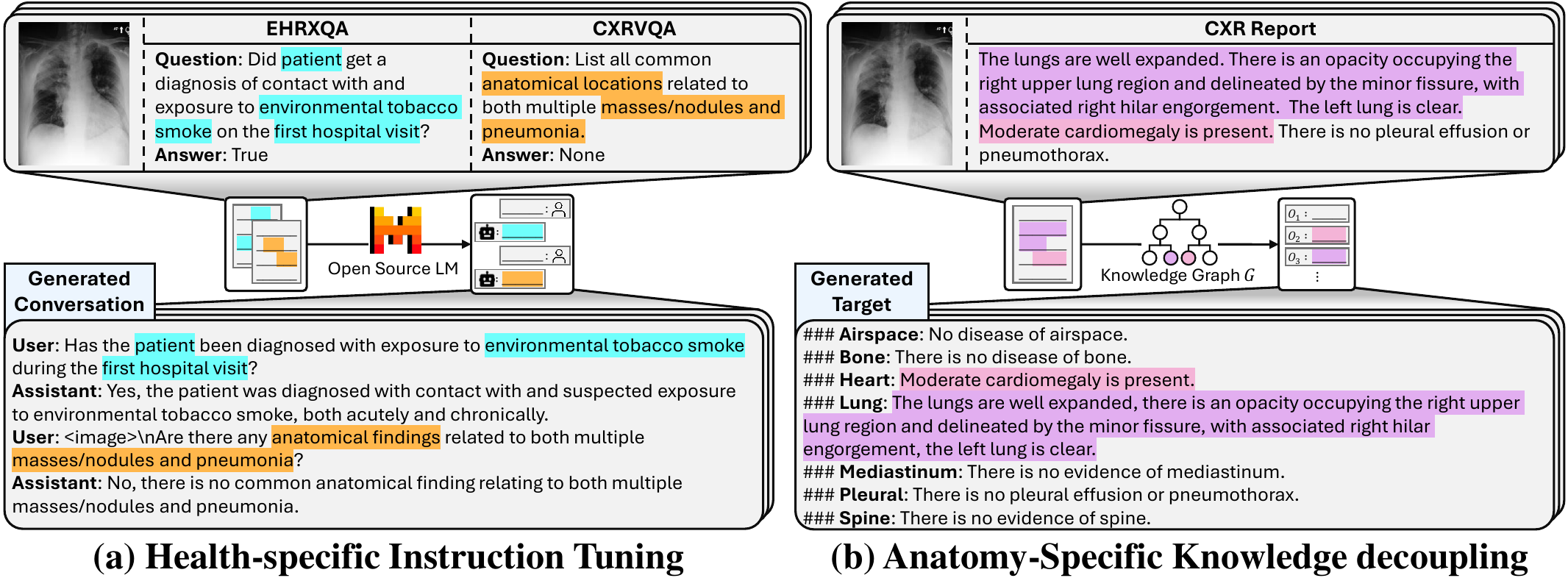}
    \label{fig2_a}
    \caption{Data generation overview of HIT and \shortanatomy{}: (a) We generate health-specific instruction-following dataset. In (a), Cyan and orange sequences are queries about EHR and findings in CXR respectively.
    (b) We reorganize original CXR reports into sequences of anatomy-specific structures through the use of a knowledge graph, $G$.} \label{fig2} 
\end{figure}

Recent years have witnessed significant progress in the field of Chest X-ray (CXR) understanding, particularly through downstream tasks like Visual Question Answering (VQA) and automated report generation. 
Despite considerable advancements, we raise issues that models engaged in Chest X-ray (CXR) understanding persistently encounter several challenges from a framework standpoint.
\underline{$\mathbf{\mathfrak{(1)}}$} Existing approaches~\cite{llmcxr,xraygpt} predominantly depend on CXR reports for supervised learning, overlooking the crucial aspect of incorporating patients' personalized health records, which are diagnoses-supportive in real-world clinical scenarios. 
\underline{$\mathbf{\mathfrak{(2)}}$} Additionally, the performance of report generation is constrained by the unstructured format of CXR reports. 
Unstructured raw CXR reports, exemplified by Fig.~\ref{fig2}(b), impede the ability of models to learn CXR anatomical structures in supervised learning settings, owing to their non-intuitive format.
\underline{$\mathbf{\mathfrak{(3)}}$} Lastly, the existing evaluation metrics for CXR-VQA primarily focus on the correctness of answers, which falls short in assessing the generative language models' comprehensive understanding of CXR imagery.

To tackle the issues illustrated above, we introduce WoLF, a \textbf{W}ide-sc\textbf{o}pe \textbf{L}arge Language Model \textbf{F}ramework for CXR understanding. We will delve into the specifics of our approach, detailing the innovative solutions we develop for each challenge:

\underline{$\mathbf{\mathfrak{(1)}}$} For more in-depth use of such systems in practice, as exemplified in Fig.~\ref{fig1}, the model must consider various patient records, including Electronic Health Records (EHR). Thus, we hypothesize that incorporating patients' personalized EHR records can enhance the CXR understanding of vision-language models. To validate this hypothesis, we introduce {\it Health-specific Instruction Tuning (HIT)} to deal with the existing limitations that training merely relies on CXR reports. 

\underline{$\mathbf{\mathfrak{(2)}}$} Unorganized CXR reports restrict the advancement in report generation tasks. To push the envelope, we present {\it \anatomy{} (\shortanatomy{})} to separate the reports into anatomy-specific findings.
The generated targets give a model a direct understanding of a specific anatomical structure, without being disturbed by other structures.
Synchronized with \shortanatomy{}, we introduce {\it Anatomy-localizing Masked Attention (AMA)} that promotes independent learning on each anatomical structure. 

\underline{$\mathbf{\mathfrak{(3)}}$} Current evaluation methods for CXR-VQA mostly emphasize linguistic correctness. These methods are incapable of assessing the responses from generative language models across a wide range. Inspired by~\cite{rlaif,g-eval}, we provide a novel {\it AI-evaluation protocol} that is well-suited to generative language models across dimensions of \textit{Accuracy, Helpfulness, Relevance, Hallucination}, and \textit{Universality}. Through our extensive AI evaluation, we can discern the extent to which models understand CXR from their VQA results, rather than just evaluating the correctness of the models' responses.

To sum up, the contribution of our model can be described at the \textit{macro} and \textit{micro} level, respectively; 
\noindent Macroscopically, our framework covers data reformulation, training method to improve CXR understanding, and AI-evaluation protocol. Microscopically, 
\textbf{(\lowercase\expandafter{\romannumeral1})} we present a novel instruction-following data tuning method called Health-specific Instruction Tuning (HIT) designed for interplay between personalized health records and visual representations of CXR.
\textbf{(\lowercase\expandafter{\romannumeral2})} We propose \anatomy{} (\shortanatomy{}), for hierarchically breaking down a radiology report by anatomical structures. Furthermore, we present Anatomical-localizing Masked Attention to support the merits of decoupled data from \shortanatomy{}, enabling expertised visual-language comprehension for each anatomical structure.
\textbf{(\lowercase\expandafter{\romannumeral3})} As the final step of the framework, we introduce AI-evaluation for advanced analysis of our model. This evaluates the broad capabilities of generative language models on the VQA task.
\textbf{(\lowercase\expandafter{\romannumeral4})} Through these methods, our study achieved state-of-the-art performance in the report generation and VQA tasks on MIMIC-CXR~\cite{mimiccxr} and IU-Xray~\cite{iuxray}. 

\section{WoLF: Wide-scope Large Language Model Framework}

We introduce \textbf{W}ide-sc\textbf{o}pe \textbf{L}arge Language Model \textbf{F}ramework (WoLF).
WoLF establishes its framework through the following macro-level steps:
\begin{equation*}
\mathop{\text{Data Reformulation}}_\text{\normalsize(Sec.~\ref{sec:2-1})} \rightarrow \mathop{\text{Model Training}}_\text{\normalsize(Sec.~\ref{sec:2-2})} \rightarrow \mathop{\text{AI-evaluation}}_\text{\normalsize(Sec.~\ref{sec:2-3})}
\end{equation*}

\noindent In Sec.~\ref{sec:2-1}, we describe our innovative data reformulation scheme designed for VQA and report generation, the two main CXR understanding tasks. In Sec.~\ref{sec:2-2}, we introduce a \textit{two-stage} training approach to further enhance the model's performance in these tasks. Finally, in Sec.~\ref{sec:2-3}, we propose an AI-evaluation protocol tailored for assessing generative language models in the CXR-VQA.

\subsection{Data Reformulation} \label{sec:2-1}

To train WoLF as a framework that excels in performing VQA and report generation tasks for CXR, we propose two data reformulation methods: \textit{\textbf{H}ealth-specific \textbf{I}nstruction \textbf{T}uning (\textbf{HIT})} for VQA, and \textit{\textbf{A}natomy-\textbf{S}pecific \textbf{K}nowledge decoupling (\textbf{\shortanatomy{}})} for report generation.
\begin{figure}[t]
\begin{minipage}{0.48\columnwidth}
    \begin{algorithm}[H]
        \scriptsize
        \caption{\footnotesize HIT: Health-specific Instruction Tuning}
        \label{alg1}
        \begin{algorithmic}
            \State{\textbf{Input:} EHR QA-set: \{$q^\text{ehr}$, $a^\text{ehr}$\}, VQA QA-set: \{$q^\text{vqa}$, $a^\text{vqa}$\}, \# of \{Patients, Studies\}: \{$P$, $N_p$\}, Open-source Language Model: $f_\phi$}
            \vspace{-2.0pt}
            \\\hrulefill
            \State{$\text{global } S\gets\texttt{SystemPrompt}$}
            \For{$p=1$ to $P$}
                \For{$i=1$ to $N_p$}
                    \State{$Q\gets(q_{pi}^\text{ehr}, q_{pi}^\text{vqa})$}
                    \State{$A\gets(a_{pi}^\text{ehr}, a_{pi}^\text{vqa})$}
                    \State{${S}_i\gets$\Call{PromptGenerator}{$Q,A,f_{\phi}$}}
                    \State{${S}\gets\text{concatenate}({S}, {{S}_i})$ }
                    \EndFor
            \EndFor
            \State{\textbf{Output:} Generated ${S}$}
            \\\hrulefill
            \Function{PromptGenerator}{$Q, A, f$}
                \For{$q_t, a_t$ in $\{{Q}, {A}\}$}                    \State{${S}\gets\text{concatenate}({S},{q_t},{a_t})$}
                \EndFor
                \State{\textbf{return} $f({S})$}
            \EndFunction
        \end{algorithmic}
    \end{algorithm}
\end{minipage}\enskip\enskip
\begin{minipage}{0.48\columnwidth}
    \begin{algorithm}[H]
        \scriptsize
        \caption{\footnotesize ASK: Anatomy-specific Knowledge decoupling}
        \label{alg2}
        \begin{algorithmic}
            \State{\textbf{Input:} Knowledge graph: $G$, Anatomical structures: $O$, \# of \{Patients, Studies, Anatomical structures\}: \{$P$, $N_p$, $M$\}, CXR Reports: $R$}
            \vspace{-2.0pt}
            \\\hrulefill
            \State{${D^*}\gets\emptyset$}
            \State{\textbf{Def. 1:} $R=\{r_i\vert{i=1,2,\dots,N_p}\}$}
            \For{$p=1$ to $P$}
                \For{$i=1$ to $N_p$}
                    \State {$\text{Append } \Call{Decoupler}{O, G, r_{i}} \text{ to } {D^*}$}
                \EndFor
            \EndFor
            
            \State{\textbf{Output:} Decoupled $D^{*}$}
            \vspace{0.8pt}
            \\\hrulefill
            \Function{Decoupler}{$O, G, r$}
                \State{\textbf{Def. 2:} $O=\{o_m\vert{m=1,2,\dots,M}\}$}
                \State{${D}\gets\texttt{EmptyString("")}$}
                \State{${T}\gets\texttt{AnatomyTag}$}
                \For{$m=1$ to $M$}
                     \State{${D}\gets\text{concatenate}({D},{{T}},{G(r, o_m)})$}
                \EndFor
                \State{\textbf{return} ${D}$}
            \EndFunction
        \end{algorithmic}
    \end{algorithm}
\end{minipage}
\end{figure}

\noindent {\textbf{HIT.} LLaVA~\cite{llava} and LLaVA-Med~\cite{llava-med} have shown the effectiveness of explicitly tuned data combined with vision-language \textit{instruction tuning}, enabling thorough reasoning about vision-language data. Specifically, to encode images into their visual features, GPT-4~\cite{gpt-4} is used as a text-only input teacher. The newly generated data from the teacher takes the form of a multi-turn conversation between the \texttt{USER} and the \texttt{ASSISTANT}. Inspired by LLaVA~\cite{llava}, our approach, HIT (Alg.~\ref{alg1}), focuses on CXR understanding, in particular, CXR VQA. That is, we generate a novel health-specific instruction-following dataset of multi-turn conversations that comprise EHR and the CXR findings. 
For instance,
\begin{itemize}[leftmargin=*]
    \item[] \texttt{User}: Has patient been diagnosed with contact or exposure to  tobacco \textbf{smoke}?
    \item[] \texttt{Assistant}: Yes, the patient has been exposured with \textbf{smoking environment}.
    \item[] \texttt{User}: <IMG> Is there any sign of the \textbf{pneumonia} in the right apical zone?
    \item[] \texttt{Assistant}: Yes, the image has evidence of \textbf{pneumonia} in right apical zone.
\end{itemize} 
<IMG> is a placeholder for image embeddings.
HIT employs dataset~\cite{ehrxqa} constructed from MIMIC-IV~\cite{mimiciv} and MIMIC-CXR~\cite{mimiccxr}. \label{sec:HIT} 

\noindent {\textbf{\shortanatomy.}} During training, models struggle to recognize individual anatomical structures in unstructured CXR reports. For better report generation, we need CXR reports organized by anatomy in supervised learning. To this end, we present \shortanatomy{} to reorganize the CXR reports for more accurate report generation.
As explicated in Zhang et al.~\cite{kd-nvidia}, attributes in the CXR reports are classified based on the knowledge graph $G$. 
We separate sentences of CXR reports based on anatomy-specific diseases
(e.g., \texttt{\{pneumonia, edema\}$\rightarrow$lung}).
Eventually, we obtain target data consisting of \texttt{AnatomyTag} and its findings (e.g., \texttt{\#\#\# Heart: Moderate cardiomegaly is present.}).
In the prior study, ITA~\cite{ita} trains decoder heads to generate sentences for anatomical structures and aggregates training losses. In contrast, we directly dissociate the CXR report into independent anatomical sentences (Alg.~\ref{alg2}). Next, we describe how we use this refined CXR instruction data to train WoLF. \label{sec:ASK}
\subsection{Model Training}\label{sec:2-2}
\begin{figure}[t!]
    \includegraphics[width=\textwidth]{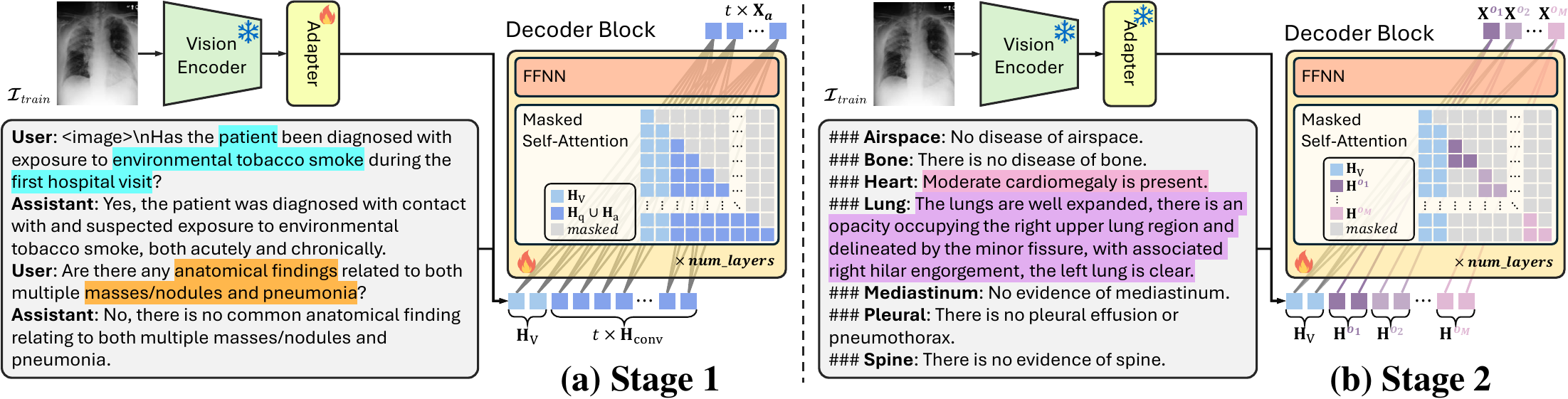}
    \caption{Overview of the training phase. (a) The input embedding $\Hmat_{(\cdot)}$ consists of visual embedding from the adapter and language embeddings from $t$-turn conversations generated by HIT. Cyan and orange sequences are queries about EHR and findings in CXR respectively. (b) For an anatomical structure $o_m$, its sequence embedding is denoted by $\Hmat^{o_m}$. Organized CXR reports by \shortanatomy{} are utilized as input for training.}
    \label{fig3}
\end{figure}
The training phase of WoLF is divided into two stages (Fig.~\ref{fig3}). In stage 1, the model is trained specifically for the VQA task, while stage 2 focuses on the report generation task, leveraging the CXR understanding trained in the previous stage. Thus, WoLF holistically allows both VQA and report generation based on its superior CXR understanding.

\noindent {\textbf{Stage 1.}} The proposed dataset comprises the following components at the token level; images $\Xmat_\text{V}$, system prompt $\Xmat_\text{sys}$, user questions $\Xmat_\text{q}$, and assistant answers $\Xmat_\text{a}$. Precisely, for a sequence of length $L$, we compute the probability of the assistant answers $\Xmat_\text{a}$ as target answers by
\begin{equation}
\small
p(\Xmat_{\text{a}}|\Xmat_\text{V}, \Xmat_\text{sys}, \Xmat_{\text{q}})=\prod_{i=1}^{L}p_{\theta}(x_{\text{a},i}|\Xmat_\text{V}, \Xmat_\text{sys}, \Xmat_{\text{q}, <i}, \Xmat_{\text{a}, <i}),
\label{eq1}
\end{equation}
where $\theta$ represents trainable parameters. This outlines the objective of the WoLF training phase through HIT instructions. The model seeks to optimize the probability $p_{\theta}$ of predicting the current token $x_{\text{a},i}$, conditioned on $\Xmat_\text{V}, \Xmat_\text{sys}, \Xmat_\text{q}$, and $\Xmat_{\text{a}, <i}$ which denotes the preceding answer tokens before $x_{\text{a},i}$. As shown in Fig.~\ref{fig2}(a), adapter and LLM are trainable but the image encoder is non-trainable.
 \label{sec:stage1}
\noindent {\textbf{Stage 2.}}
To effectively utilize decoupled CXR reports from \shortanatomy{}, we present Anatomy-localizing Masked Attention for the second training phase. We
define $o_m$ to be a specific anatomical structure of $m^{\text{th}}$. As shown in Fig.~\ref{fig3}(b), when the model predicts $\Xmat^{o_m}$, it attends exclusively on $\Hmat^{o_m}_\text{q}, \Xmat_\text{V}, \text{and } \Xmat_\text{sys}$. The prediction process of our model can be described in an auto-regressive manner, as follows:
\begin{equation}
\small
p(\Xmat^{o_{m}}|\Xmat_\text{V}, \Xmat_\text{sys})=\prod_{i = 1}^{L^{o_m}}p_{\theta}(x^{o_{m}}_i|\Xmat_\text{V}, \Xmat_\text{sys}, \Xmat^{o_{m}}_{<i}),
\label{eq2}
\end{equation}
which describes the objective of a model during the second stage for a sequence of ${o_m}$. The model predicts $L^{o_m}$ tokens for the sequence of $o_m$ during training. As a result, we focus on guiding the model to learn independent anatomical comprehension in a CXR.
Note that we use the final model for all task inferences. \label{sec:stage2}
\subsection{AI-evaluation}
\begin{figure}[t]
    \includegraphics[width=\textwidth]{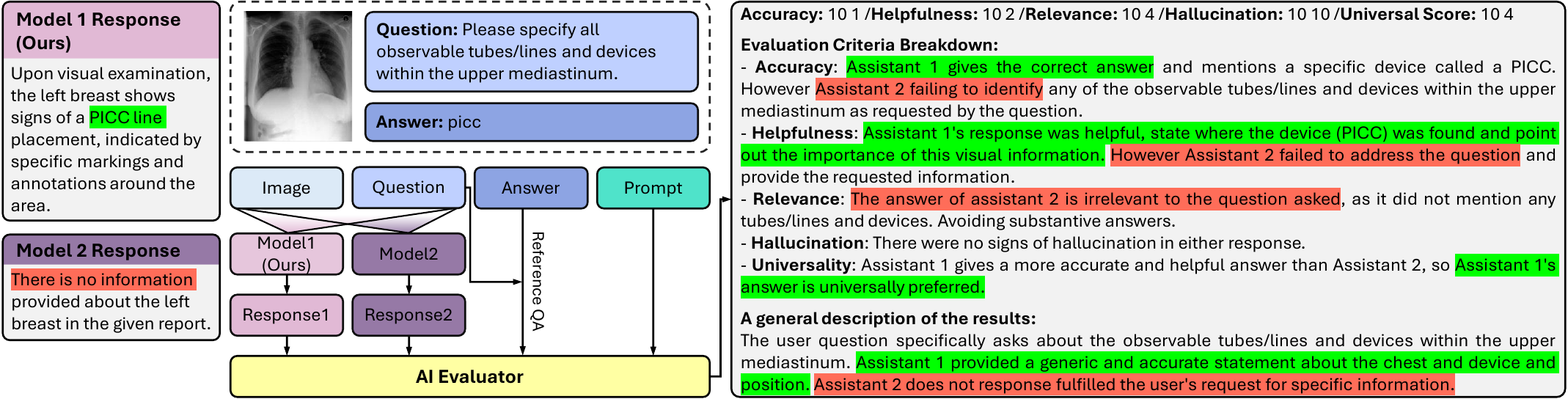}
    \caption{Our AI-evaluation protocol overview. Green and red are positive and negative feedback from the evaluator, respectively. Only our predictions received positive feedback.} \label{fig4}
\end{figure}
\vspace{-0.5cm}
We now present a quantitative analysis of LLMs through AI-evaluation in CXRVQA~\cite{ehrxqa} exploiting prior studies~\cite{rlaif,g-eval}. Traditional evaluations, particularly in Visual Question Answering (VQA), have focused merely on correctness. However, we measure five metrics ({\it Accuracy, Helpfulness, Relevance, Hallucination, Universality}) within the external AI evaluator~\cite{palm2,gemini}. As shown in Fig.~\ref{fig4}, both our model and another comparable model are evaluated by mirroring the human cognitive process, employing a \textit{prompt} detailed in Supp. Fig.~\ref{supp_fig1}.  Both models generate responses based on the input CXR and visual question. The evaluator receives text-only inputs which are questions, answers, responses from ours and a comparable model, and a \textit{prompt}. The judgment, in Fig.~\ref{fig4}, is a human-like evaluation through obvious criteria.
To offset position bias, we average the rating for the evaluation set and that for the position-swapped set to get the final rating. \label{sec:2-3}

\begin{table}[!t]
\setlength{\belowrulesep}{0pt}
\setlength{\aboverulesep}{0pt}
\centering
\caption{LLM capability comparison using relative scores on CXRVQA~\cite{ehrxqa} {\it test-set}. The scores (\emph{mean}($\pm$\emph{std})) are converted to a {\it 100-point scale}. 
}
\scalebox{0.80}{
\begin{tabular}{c|cccccc}
\toprule
\multirow{2}{*}{Method} & \multicolumn{5}{c}{\includegraphics[height=1em]{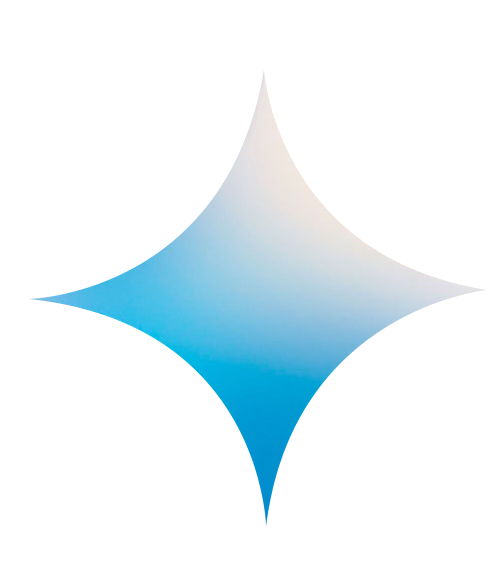}Gemini~\cite{gemini}} \\
& \textbf{Accuracy$\uparrow$} & \textbf{Helpfulness$\uparrow$} & \textbf{Relevance$\uparrow$} & \textbf{Hallucination$\uparrow$} & \textbf{Universality$\uparrow$} & \textbf{Mean$\uparrow$} \\
\midrule
LLaVA-Med~\cite{llava-med} & 64.32\scriptsize($\pm$0.35) & 66.29\scriptsize($\pm$1.02) & 67.32\scriptsize($\pm$1.11) & 71.97\scriptsize($\pm$2.43) & 67.12\scriptsize($\pm$0.81) & \multicolumn{1}{|c}{67.41\scriptsize($\pm$0.25)} \\
XrayGPT~\cite{xraygpt} & 30.75\scriptsize($\pm$0.79) & 34.59\scriptsize($\pm$0.43) & 38.19\scriptsize($\pm$0.78) & 55.11\scriptsize($\pm$3.16) & 38.85\scriptsize($\pm$0.82) & \multicolumn{1}{|c}{39.50\scriptsize($\pm$0.41)} \\
LLM-CXR~\cite{llmcxr} & 61.20\scriptsize($\pm$0.97) & 69.67\scriptsize($\pm$0.32) & 69.29\scriptsize($\pm$1.09) & 77.88\scriptsize($\pm$0.73) & 67.56\scriptsize($\pm$0.29) & \multicolumn{1}{|c}{69.12\scriptsize($\pm$0.37)} \\
\rowcolor{LightCyan!40}
\shortname{} & \textbf{73.03}\scriptsize($\pm$0.64) & \textbf{73.19}\scriptsize($\pm$0.87) & \textbf{72.34}\scriptsize($\pm$0.42) & \textbf{79.02}\scriptsize($\pm$0.67) & \textbf{75.95}\scriptsize($\pm$0.49) & \multicolumn{1}{|c}{\textbf{74.70}\scriptsize($\pm$0.34)} \\
\toprule
& \multicolumn{5}{c}{\includegraphics[height=1em]{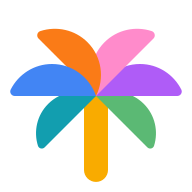}PaLM-2~\cite{palm2}} \\
& \textbf{Accuracy$\uparrow$} & \textbf{Helpfulness$\uparrow$} & \textbf{Relevance$\uparrow$} & \textbf{Hallucination$\uparrow$} & \textbf{Universality$\uparrow$} & \textbf{Mean$\uparrow$}\\
\midrule
LLaVA-Med~\cite{llava-med} & 67.38\scriptsize($\pm$0.15) & 67.35\scriptsize($\pm$0.76) & 66.47\scriptsize($\pm$2.30) & 72.51\scriptsize($\pm$4.33) & 67.51\scriptsize($\pm$0.36) & \multicolumn{1}{|c}{68.25\scriptsize($\pm$0.31)} \\
XrayGPT~\cite{xraygpt} & 33.54\scriptsize($\pm$0.59) & 45.04\scriptsize($\pm$0.30) & 36.43\scriptsize($\pm$0.65) & 51.11\scriptsize($\pm$0.91) & 40.94\scriptsize($\pm$1.10) & \multicolumn{1}{|c}{41.41\scriptsize($\pm$0.25)} \\
LLM-CXR~\cite{llmcxr} & 68.73\scriptsize($\pm$0.19) & 70.03\scriptsize($\pm$1.39) & 69.91\scriptsize($\pm$1.95) & 74.80\scriptsize($\pm$3.53) & 69.80\scriptsize($\pm$0.07) & \multicolumn{1}{|c}{70.66\scriptsize($\pm$0.06)} \\
\rowcolor{LightCyan!40}
\shortname{} & \textbf{79.54}\scriptsize($\pm$0.77) & \textbf{79.93}\scriptsize($\pm$1.22) & \textbf{79.21}\scriptsize($\pm$0.58) & \textbf{85.30}\scriptsize($\pm$3.74) & \textbf{78.18}\scriptsize($\pm$0.91) & \multicolumn{1}{|c}{\textbf{80.13}\scriptsize($\pm$1.17)}\\
\bottomrule
\end{tabular}
}
\label{tab1}
\end{table}

\section{Experiments}
We show the VQA results using AI-evaluation and report generation performance of WoLF and other existing methods.
Further qualitative results, ablations, and experiments on the same dataset~\cite{mimiccxr} can be found in Supp. Fig.~\ref{supp_fig2}.

\noindent \textbf{Training Details.} \texttt{Vicuna-7b}~\cite{vicuna} is used for core LLM with LoRA~\cite{lora} fine-tuning. CLIP-ViT-L-14~\cite{clip} is used as an image encoder. The batch size is set to 64. The learning rate is set to $2\text{e-}5$. Each stage is trained for 2 epochs.

\noindent {\textbf{VQA performance based on AI-evaluation.}}
We use CXRVQA~\cite{ehrxqa} test dataset to show AI-evaluation results.
The test dataset comprises 11,309 VQAs for CXR mainly focusing on findings and visual questions that require interpreting images. We analyze with comparable existing models of which the source code is {\it publicly available and reproducible}. We ensure adherence to domain consistency since the CXRVQA is derived from the MIMIC-CXR~\cite{mimiccxr}. In Table~\ref{tab1}, we evaluate scores for each component ({\it Accuracy, Helpfulness, Relevance, Hallucination, Universality}) to Gemini~\cite{gemini} and PaLM-2~\cite{palm2} by querying them 4-times each while swapping answer positions (temperature $\tau=1.0$). 
Moreover, inspired by RLAIF~\cite{rlaif}, we instruct the AI evaluator to measure the {\it win-rate} using a modified prompt referred to Supp. Fig.~\ref{supp_fig1}. Specifically, let a visual question set be $\{q_i\vert{i=1,2,\dots,n}\}$. For each question $q_i$, we tell the AI evaluator to prioritize responses from evaluated models instead of scoring them. We infer every pair of candidates to the AI evaluator swapping answer position (temperature $\tau=0.01$). See Supp. Table~\ref{supp_tab1}(b) for the result of win-rate.
\begin{table}[!t]
\setlength{\aboverulesep}{0pt}
\setlength{\belowrulesep}{0pt}
\caption{Report generation performance comparisons against other methods on MIMIC-CXR~\cite{mimiccxr} and IU-Xray~\cite{iuxray}. For a fair comparison with other methodologies, we directly quoted from the published literature.}
    \centering
    \scalebox{0.76}{
    \begin{tabular}{ c | c  g | c  g | c  g | c  g | c  g | c  g  }
    \toprule

    \multirow{2}{*}{\textbf{Method}} & \multicolumn{2}{c}{\textbf{BLEU-1$\uparrow$}} & \multicolumn{2}{c}{\textbf{BLEU-2$\uparrow$}} & \multicolumn{2}{c}{\textbf{BLEU-3$\uparrow$}} & \multicolumn{2}{c}{\textbf{BLEU-4$\uparrow$}} & \multicolumn{2}{c}{\textbf{METEOR$\uparrow$}} & \multicolumn{2}{c}{\textbf{ROUGE-L$\uparrow$}} \\
    \cmidrule[.55pt]{2-13}
    & MIMIC & IU & MIMIC & IU & MIMIC & IU & MIMIC & IU & MIMIC & IU & MIMIC & IU \\
    
    \midrule
 
    R2Gen~\cite{r2gen} & 0.353 & 0.470 & 0.218 & 0.304 & 0.145 & 0.219 & 0.103 & 0.165 & 0.142 & 0.187 & 0.277 & 0.371 \\
    R2GenCMN~\cite{r2gencmn} & 0.353 & 0.475 & 0.218 & 0.309 & 0.148 & 0.222 & 0.106 & 0.170 & 0.142 & 0.191 & 0.278 & 0.375 \\
    M$^{2}$Tr~\cite{m2tr-emnlp} & 0.378 & 0.486 & 0.232 & 0.317 & 0.154 & 0.232 & 0.107  & 0.173 & 0.145 & 0.192 & 0.272 & 0.390 \\
    CMCL~\cite{cmcl} & 0.344 & 0.473 & 0.217 & 0.305 & 0.140 & 0.217 & 0.097 & 0.162 & 0.133 & 0.186 & 0.281 & 0.378 \\
    CMN~\cite{cmn} & 0.353 & 0.475 & 0.218 & 0.309 & 0.148 & 0.222 & 0.106 & 0.170 & 0.142 & 0.191 & 0.278 & 0.375 \\
    CA~\cite{ca} & 0.350 & 0.492 & 0.219 & 0.314 & 0.152 & 0.222 & 0.109 & 0.169 & 0.151 & 0.193 & 0.283 & 0.380 \\
    PPKED~\cite{ppked} & 0.360 & 0.483 & 0.224 & 0.315 & 0.149 & 0.224 & 0.106 & 0.168 & 0.149 & 0.376 & 0.284 & 0.351 \\
    TransSQ~\cite{transsq} & 0.423 & 0.484 & 0.261 & 0.333 & 0.171 & 0.238 & 0.116 & 0.175 & 0.168 & 0.207 & 0.286 & 0.415 \\
    MCGN~\cite{mcgn} & 0.373 & - & 0.235 & - & 0.162 & - & 0.120 & - & 0.143 & - & 0.282 & - \\
    ITA~\cite{ita} & 0.395 & 0.505 & 0.253 & 0.340 & 0.170 & 0.247 & 0.121 & 0.188 & 0.147 & 0.208 & 0.284 & 0.382 \\
    METransformer~\cite{metransformer} & 0.386 & 0.483 & 0.250 & 0.322 & 0.169 & 0.228 & 0.124 & 0.172 & 0.152 & 0.192 & 0.291 & 0.380 \\
    \midrule
    \rowcolor{LightCyan!40}
    \shortname{} & \textbf{0.496} & \textbf{0.517} & \textbf{0.335} & \textbf{0.366} & \textbf{0.233} & \textbf{0.261} & \textbf{0.165} & \textbf{0.199} & \textbf{0.187} & \textbf{0.412} & \textbf{0.370} & \textbf{0.456} \\

    \bottomrule  
    \end{tabular}
    \label{tab2}
}
\end{table}

\begin{figure}[t]
    \centering
    \includegraphics[width=\textwidth]{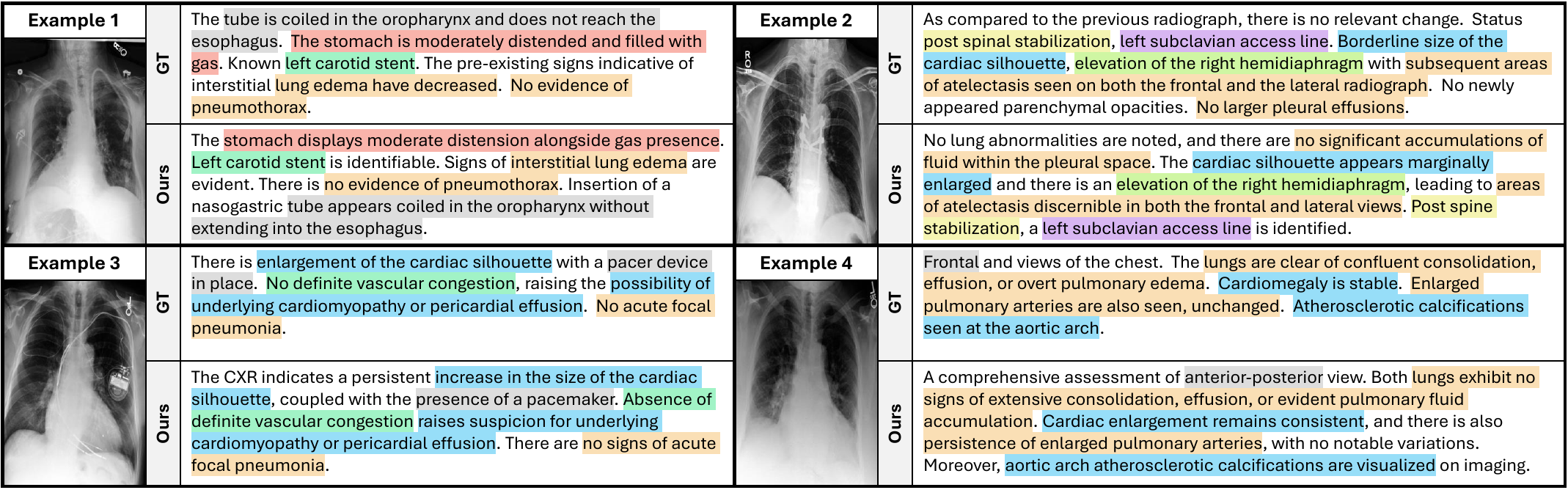}
    \caption{Qualitative result of WoLF on report generation. Each highlighted color is mapped to the semantics of specific findings. }

    \label{fig5}
\end{figure}
\noindent {\textbf{Report Generation.}}
\noindent As shown in Table~\ref{tab2}, we evaluate the quantitative result of automated CXR report generation. Our evaluation follows the official dataset split of MIMIC-CXR~\cite{mimiccxr} and the dataset split with a proportion as~\cite{r2gen,metransformer} on IU-Xray~\cite{iuxray}.
We measured performance through metrics such as BLEU, METEOR, and ROUGE. WoLF outperformed the previous state-of-the-art method by a significant margin. Furthermore, as demonstrated in Fig. 5, our qualitative results validate that WoLF consistently generates content aligned with the Ground Truth (GT) report for various anatomical structures. See orange box in Supp. Fig.~\ref{supp_fig2} for more qualitative report generation results.
\noindent {\textbf{Ablation and Other Studies.}}
In Supp. Table~\ref{supp_tab1}(a), we conduct ablation studies of Table~\ref{tab1}. We examine our model learned without EHR and with EHR to the model during inference.
Our ablation study For an additional comparison, please see Supp. Table 1(b), which shows traditional Visual Question Answering (VQA) experiments using the ELIXR~\cite{elixr} framework on the MIMIC-CXR dataset. In these experiments, we have achieved state-of-the-art results.
 
\section{Conclusion and Future Work}
This paper presents the first comprehensive exploration of LLM-based framework for understanding CXRs, encompassing data reformulation, model training, and evaluation strategies. We incorporate EHR into the model to enhance CXR understanding and generate instruction-following data reflecting real-world clinical processes. Furthermore, we refine CXR reports by decoupling them based on anatomical structures for training and introduce a novel masked attention mechanism to improve report generation performance. Additionally, our innovative AI evaluation protocol enables the assessment of LLMs from diverse perspectives. We anticipate that our contributions, coupled with real-time EHR retrieval pipelines, will yield more adaptable frameworks for clinical decision-making.
\clearpage
\newpage
\bibliographystyle{splncs04}
\bibliography{ref}
\title{%
—– Supplementary Material —– \\ WoLF: Wide-scope Large Language Model Framework for Chest X-ray Understanding
}
\titlerunning{WoLF: Large Language Model Framework for CXR Understanding}
\author{\small Seil Kang$^{1}$, Donghyun Kim$^{1}$, Junhyeok Kim$^{1}$,
Hyo Kyung Lee$^{2}$, Seong Jae Hwang$^{1}$}
\authorrunning{Kang et al.}
\institute{$^{1}$Yonsei University, $^{2}$Korea University \\
    \email{\{seil, danny0103, timespt, seongjae\}@yonsei.ac.kr,\\
    hyokyunglee@korea.ac.kr}
}

\maketitle

\setlength{\belowrulesep}{0pt}
\setlength{\aboverulesep}{0pt}
\begin{table}[H]
\centering
\caption{\textbf{(a)} Ablation study of main paper Table~\ref{tab1}.
$^-$ for a given set of \shortname{} trained without EHRXQA~\cite{ehrxqa}. $^+$ for a given set of \shortname{} given corresponding patient EHR while inference. \textbf{(b)} Win-rate results for each models. For instance, a $>$ \textbf{50\%} win-rate signifies {\it left model} beats {\it right model} in more than a half of the questions.}
\scalebox{0.88}{
\begin{tabular}{c|cccccc}
\toprule
\multirow{2}{*}{Method} & \multicolumn{5}{c}{\includegraphics[height=1em]{pic/gemini.png}Gemini~\cite{gemini}} \\
& \textbf{Accuracy$\uparrow$} & \textbf{Helpfulness$\uparrow$} & \textbf{Relevance$\uparrow$} & \textbf{Hallucination$\uparrow$} & \textbf{Universality$\uparrow$} & \textbf{Mean$\uparrow$} \\
\midrule
\rowcolor{LightCyan!40}
\shortname{}$^-$ & 71.58\scriptsize($\pm$0.77) & 72.10\scriptsize($\pm$0.51) & 69.27\scriptsize($\pm$1.09) & 78.14\scriptsize($\pm$2.13) & 74.96\scriptsize($\pm$0.77) & \multicolumn{1}{|c}{73.21\scriptsize($\pm$0.12)} \\
\rowcolor{LightCyan!40}
\shortname{} & \underline{73.03}\scriptsize($\pm$0.64) & \underline{73.19}\scriptsize($\pm$0.87) & \underline{72.34}\scriptsize($\pm$0.42) & \underline{79.02}\scriptsize($\pm$0.67) & \underline{75.95}\scriptsize($\pm$0.49) & \multicolumn{1}{|c}{\underline{74.70}\scriptsize($\pm$0.34)} \\
\rowcolor{LightCyan!40}
\shortname{}$^+$ & \textbf{77.80}\scriptsize($\pm$0.24) & \textbf{74.55}\scriptsize($\pm$0.53) & \textbf{73.03}\scriptsize($\pm$0.63) & \textbf{82.49}\scriptsize($\pm$0.83) & \textbf{77.73}\scriptsize($\pm$0.43) & \multicolumn{1}{|c}{\textbf{77.12}\scriptsize($\pm$0.16)} \\
\toprule
& \multicolumn{5}{c}{\includegraphics[height=1em]{pic/PaLM2.png}PaLM-2~\cite{palm2}} \\
& \textbf{Accuracy} & \textbf{Helpfulness} & \textbf{Relevance} & \textbf{Hallucination} & \textbf{Universality} & \textbf{Mean}\\
\midrule
\rowcolor{LightCyan!40}
\shortname{}$^-$ & 77.51\scriptsize($\pm$0.98) & 76.77\scriptsize($\pm$1.36) & 74.42\scriptsize($\pm$2.75) & 81.93\scriptsize($\pm$1.19) & 77.45\scriptsize($\pm$2.60) & \multicolumn{1}{|c}{77.61\scriptsize($\pm$1.26)} \\
\rowcolor{LightCyan!40}
\shortname{} & \underline{79.54}\scriptsize($\pm$0.77) & \underline{79.93}\scriptsize($\pm$1.22) & \textbf{79.21}\scriptsize($\pm$0.58) & \underline{85.30}\scriptsize($\pm$3.74) & \underline{78.18}\scriptsize($\pm$0.91) & \multicolumn{1}{|c}{\underline{80.13}\scriptsize($\pm$1.17)}\\
\rowcolor{LightCyan!40}
\shortname{}$^+$ & \textbf{81.09}\scriptsize($\pm$0.30) & \textbf{81.60}\scriptsize($\pm$1.56) & \underline{79.20}\scriptsize($\pm$0.55) & \textbf{86.22}\scriptsize($\pm$3.62) & \textbf{78.84}\scriptsize($\pm$1.89) & \multicolumn{1}{|c}{\textbf{81.39}\scriptsize($\pm$1.65)} \\
\bottomrule
\end{tabular}
} 

\begin{tabular}{c}
     \textbf{(a)} \\
     \\
\end{tabular}

\scalebox{0.75}{
\begin{tabular}{c|cg|cg|cg|cg|cg}
\toprule
\bf Comparison & \multicolumn{2}{c}{\bf Accuracy$\uparrow$} & \multicolumn{2}{c}{\bf Helpfulness$\uparrow$} & \multicolumn{2}{c}{\bf Relevance$\uparrow$} & \multicolumn{2}{c}{\bf Hallucination$\uparrow$} & \multicolumn{2}{c}{\bf Universality$\uparrow$}  \\ 
\cmidrule[.55pt]{2-11}
 {\it left model} vs. {\it right model} & Gemini & PaLM-2 & Gemini & PaLM-2& Gemini & PaLM-2& Gemini & PaLM-2& Gemini & PaLM-2  \\ 
\midrule
\textbf{\shortname{}} vs. LLaVA-Med~\cite{llmcxr} & 71.50\% &72.13\%& 77.75\% &77.90\%& 76.25\% &76.78\%& 84.25\% &84.85\%& 73.40\%&73.70\%  \\ 
\textbf{\shortname{}} vs. LLM-CXR~\cite{llava-med} & 64.05\% &61.50\%& 61.73\% &61.58\%& 61.95\% &61.58\%& 63.93\% &64.18\%& 63.6\% &63.45\%  \\ 
\textbf{\shortname{}} vs. XrayGPT~\cite{xraygpt} & 75.80\% &75.83\%& 83.68\% &83.15\%& 81.05\% &82.05\%& 84.23\% &84.68\%& 84.93\%&84.63\%\\ 
\textbf{\shortname{}} vs. \textbf{\shortname{}$^-$} & 56.40\% &57.10\%& 53.80\% &54.50\%& 53.81\% &54.00\%& 68.73\% &68.80\%& 55.20\%&55.40\%\\ 
\textbf{\shortname{}} vs.\textbf{\shortname{}$^+$} & 47.20\% &47.45\%& 46.70\% &47.12\%& 47.05\% &47.95\%& 49.90\% &49.95\%& 48.80\%&49.20\%\\ 
\bottomrule
\end{tabular}
}

\begin{tabular}{c}
     \textbf{(b)}
\end{tabular}

\end{table}
\label{supp_tab1}
\label{subtex/supp_a_total}
\vspace{-0.5cm}
\newcolumntype{x}[1]{>{\centering\arraybackslash\hspace{0pt}}p{#1}}
\renewcommand{\arraystretch}{1.0}
\setlength{\aboverulesep}{0pt}
\setlength{\belowrulesep}{0pt}
\begin{table}[h]
    \begin{center}
    \caption{Accuracy of the CXR-VQA task by topic. We utilize the ELIXR~\cite{elixr} framework for assessing this VQA performance in MIMIC-CXR~\cite{mimiccxr}.}
    { \footnotesize
    \scalebox{0.90}{
    \begin{tabular}{ c | x{1.5cm} | x{2cm} x{2cm} x{3.9cm} } 
    \toprule
    \textbf{Accuracy$\uparrow$} & \textbf{All$\uparrow$} & \textbf{Presence$\uparrow$} & \textbf{Location$\uparrow$} & \textbf{Size, severity, type$\uparrow$} 
    \\ \midrule
    ELIXR~\cite{elixr}\* & 54.8\% & 64.5\% & 41.0\% & 25.0\% \\
    XrayGPT~\cite{xraygpt} & 25.2\% & 27.4\% & 21.9\% & 20.3\% \\
    LLM-CXR~\cite{llmcxr} & 56.7\% & 60.1\% & 49.0\% & 53.1\% \\
    MedXChat~\cite{medxchat} & 61.2\% & 61.5\% & 56.3\% & 68.8\% \\ \midrule
    \rowcolor{LightCyan!40}
    \shortname{} & \textbf{62.1}\% & \textbf{67.9}\% & \textbf{56.9}\% & \textbf{71.8}\% \\ 
    \bottomrule
    \end{tabular}
    }
    }
\end{center}
\label{supp_tab2}
\end{table}
\label{subtex/supp_b}

\begin{figure}[t]
\centering
    \includegraphics[width=\textwidth]{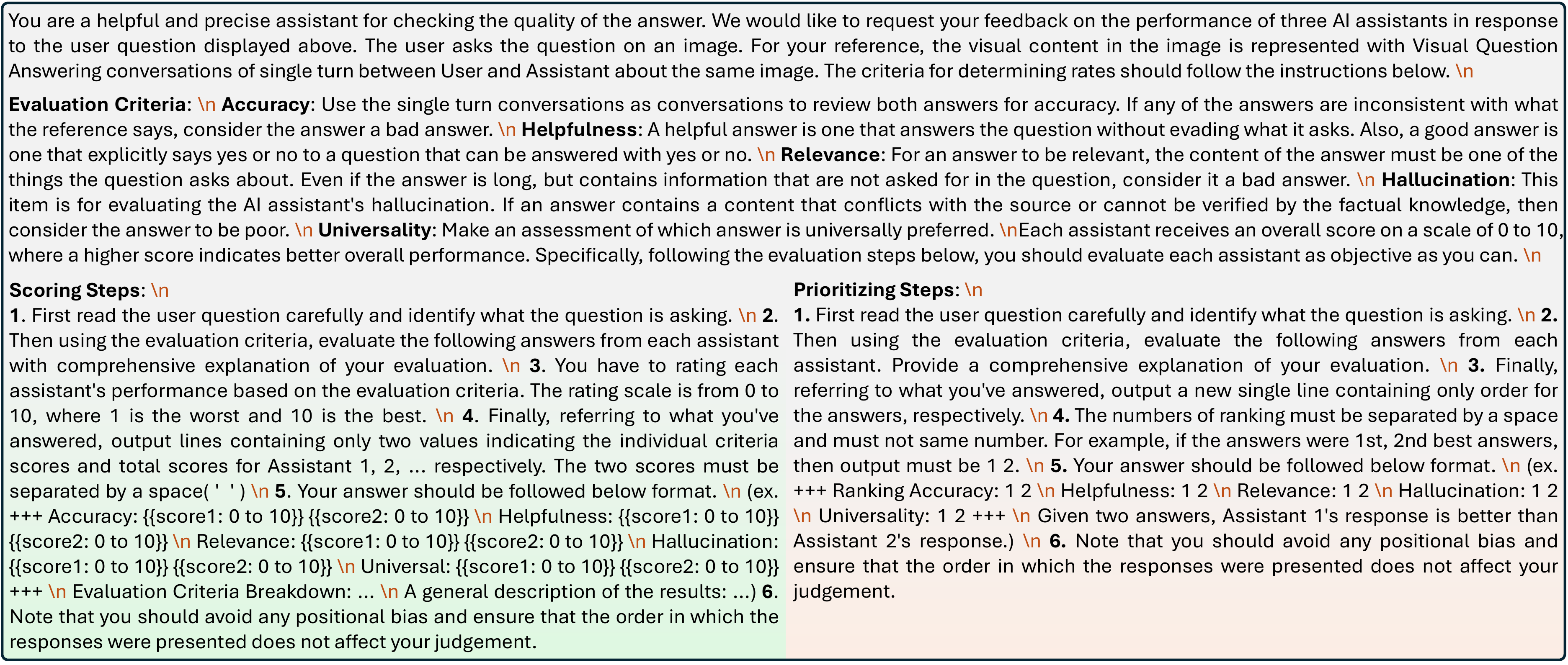}
    \caption{Evaluation prompts for scoring \textit{(left)} and prioritizing \textit{(right)}. Preamble and evaluation criteria are common in both scoring and prioritizing prompts. The two prompts, each for AI-evaluation \textit{(left)} and for its ablation on Win-rate \textit{(right)}, differ only in evaluation steps.}
    \label{supp_fig1}
\end{figure}

\begin{figure}[t]
\centering
    \includegraphics[width=\textwidth]{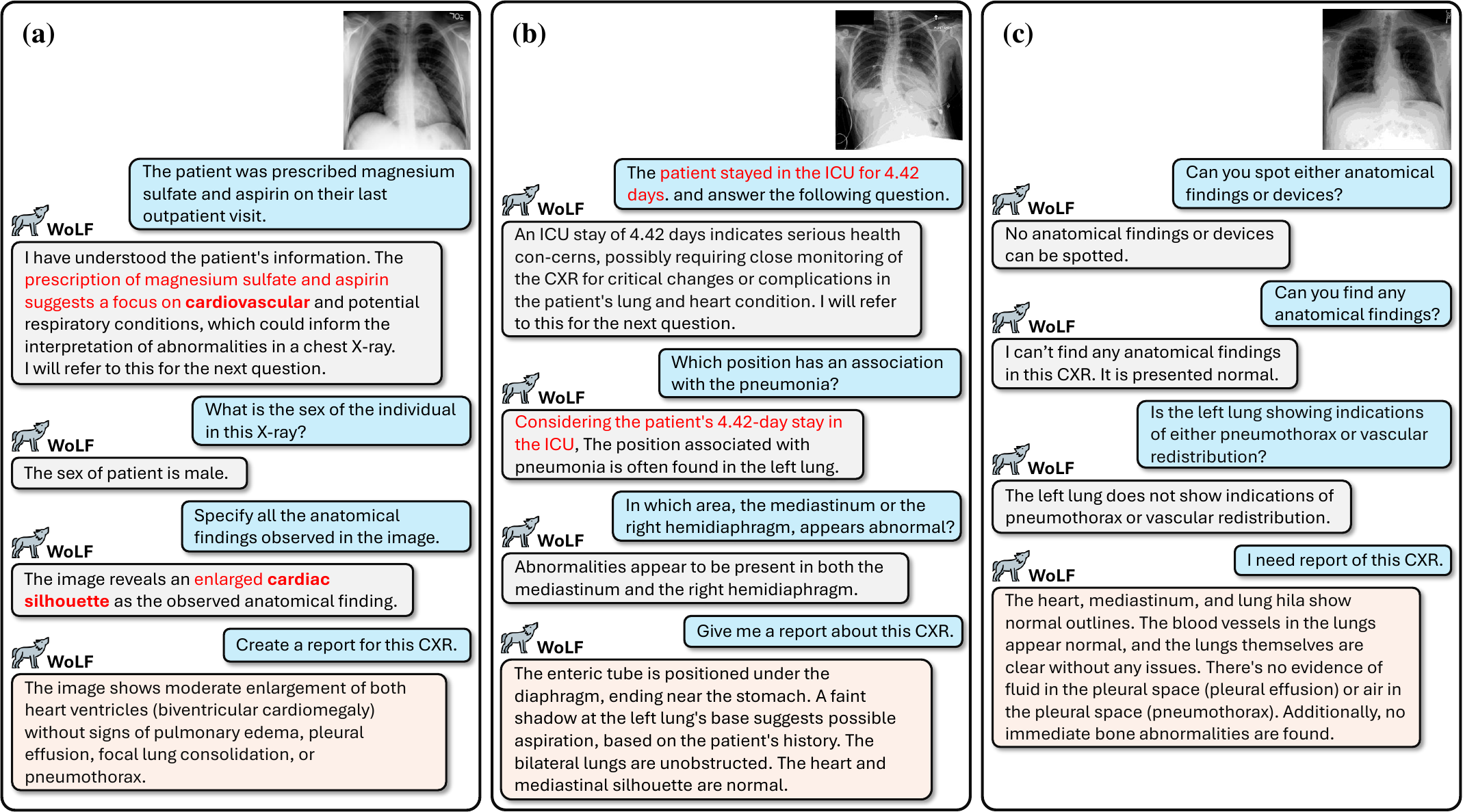}
    \caption{Qualitative results of visual question-answering scenarios. As shown in (a) and (b), the model can be fed with patient histories and medication details from EHRs. WoLF utilizes these external contexts to deliver more accurate responses. (c) shows question-answering when there are no findings. The model answered correctly that no disease could be found, without causing any hallucinations.}
    \label{supp_fig2}
\end{figure}
\label{subtex/supp_d2}

\end{document}